\documentclass[10.5pt,twocolumn,journal,letterpaper]{IEEEtran}
\usepackage{amsmath, amsthm, amssymb}
\usepackage{url,flushend,multirow,booktabs}
\usepackage{algorithm,algorithmic}   
\usepackage{graphicx}
\usepackage{bm}
\usepackage{cite}
\usepackage{subfigure}
\usepackage{float}
\usepackage[dvipsnames]{xcolor}  
\usepackage{multirow}
\usepackage[colorlinks,citecolor=Green,urlcolor=blue,bookmarks=false,hypertexnames=true]{hyperref}
\usepackage{colortbl}
\definecolor{maroon}{cmyk}{0.08,0.04,0.00,0.06}  

\renewcommand{\algorithmicrequire}{\textbf{Input:}}   
\renewcommand{\algorithmicensure}{\textbf{Output:}}  

\DeclareFixedFont{\mf}{OT1}{ptm}{m}{n}{10pt}
\DeclareFixedFont{\mfb}{OT1}{ptm}{bx}{n}{10pt}

\begin{document}
%


\title{SGCap: Decoding Semantic Group for Zero-shot Video Captioning}
%
%
%

\author{Zeyu~Pan, Ping~Li, Wenxiao~Wang
	\thanks{Z.~Pan and P.~Li are with Hangzhou Dianzi University, Hangzhou, China (e-mail: panzeyucs@hdu.edu.cn, lpcs@hdu.edu.cn). }
	\thanks{W.~Wang is with Zhejiang University, Hangzhou, China (e-mail: wenxiaowang@zju.edu.cn). P.~Li is the corresponding author.}  
}
\markboth{arXiv}
{Pan \MakeLowercase{\textit{et al.}}:~SGCap: Decoding Semantic Group for Zero-shot Video Captioning}
%

\maketitle

\begin{abstract}
Zero-shot video captioning aims to generate sentences for describing videos without training the model on video-text pairs, which remains underexplored. Existing zero-shot image captioning methods typically adopt a text-only training paradigm, where a language decoder reconstructs single-sentence embeddings obtained from CLIP. However, directly extending them to the video domain is suboptimal, as applying average pooling over all frames neglects temporal dynamics. To address this challenge, we propose a \textbf{S}emantic \textbf{G}roup \textbf{Cap}tioning (\textbf{SGCap}) method for zero-shot video captioning. In particular, it develops the Semantic Group Decoding (SGD) strategy to employ multi-frame information while explicitly modeling inter-frame temporal relationships. Furthermore, existing zero-shot captioning methods that rely on cosine similarity for sentence retrieval and reconstruct the description supervised by a single frame-level caption, fail to provide sufficient video-level supervision. To alleviate this, we introduce two key components, including the Key Sentences Selection (KSS) module and the Probability Sampling Supervision (PSS) module. The two modules construct semantically-diverse sentence groups that models temporal dynamics and guide the model to capture inter-sentence causal relationships, thereby enhancing its generalization ability to video captioning. Experimental results on several benchmarks demonstrate that SGCap significantly outperforms previous state-of-the-art zero-shot alternatives and even achieves performance competitive with fully supervised ones. Code is available at \href{https://github.com/mlvccn/SGCap_Video}{https://github.com/mlvccn/SGCap\_Video}.
\end{abstract}

\section{Introduction}
In the last decade, many contributions have been devoted to exploring ways of generating descriptive captions for visual content, including images and videos. As a fundamental multi-modal task, various solutions have been proposed by integrating Large Language Models (LLMs) with vision components \cite{li-icml2024-blip2,Liu-nips2023-llava,ye-cvpr2024-mplugowl,zhu-iclr2024-minigpt4,zhang-arxiv2023-videollama}. These approaches demonstrate strong interactive capabilities by leveraging the linguistic knowledge embedded in language models. However, they heavily depend on large-scale image-text pairs or video-text pairs to align visual information with textual concepts, making the training process costly and labor-intensive. Furthermore, when visual data are unavailable and only textual data are accessible, it becomes challenging to fine-tune these models to achieve satisfactory performance.

\begin{figure}[t]
  \centering
  \includegraphics[width=0.8\linewidth]{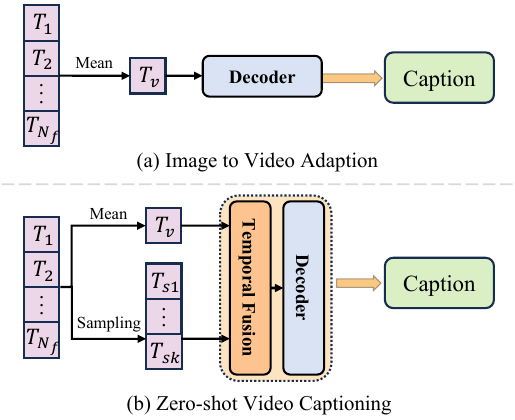}
  \caption{Comparison between previous zero-shot image captioning methods adapted for zero-shot video captioning tasks and our method.}
  \label{fig: comparison}
\end{figure}

To address the challenges of collecting visual-text pairs, recent research proposes the zero-shot captioning task. Several approaches \cite{tewel-cvpr2022-zerocap,su-arXiv2022-magic} leverage CLIP \cite{radford-icml2021-clip} to guide a pre-trained language model in generating image-correlated text by computing the cosine similarity between word and image embeddings at each step of the text generation process. Previous studies \cite{li-iclr2023-decap,wang-ijcai2023-knight,zhang-iclr24-c3,lee-emnlp2024-ifcap} have investigated a text-only training paradigm, where a decoder is trained only on textual data and subsequently used to decode CLIP text embeddings, generating corresponding image descriptions.

The task of video captioning aims to generate descriptive captions for a video clip composed of multiple frames. However, previous text-only training methods, which are effective for zero-shot image captioning task, encounter significant challenges when applied to zero-shot video captioning. Specifically, CLIP is an image model, meaning that these models fail to aggregate video frames into a single embedding. Instead, they apply average pooling across frame features, as shown in Figure \ref{fig: comparison}, which severely hinders their ability to capture temporal dynamics in videos. Although some approaches \cite{liu-aaai2024-syntic,yang-acl2023-multicapclip} incorporate additional visual cues to improve caption accuracy, they require pre-trained networks for feature extraction, leading to increased computational costs. To this end, we propose the semantic group decoding strategy to leverage multi-frame information while modeling inter-frame temporal relationships without relying on external pre-trained networks. During training, we retrieve distinctive sentences correlated with the training caption, simulating the temporal dynamics across frames. These embeddings are then fused with the training caption using a fusion module. The resulting feature is subsequently decoded by a language decoder supervised by a reconstruction loss to generate an accurate and descriptive sentence.

\begin{figure}[t]
  \centering
  \includegraphics[width=\linewidth]{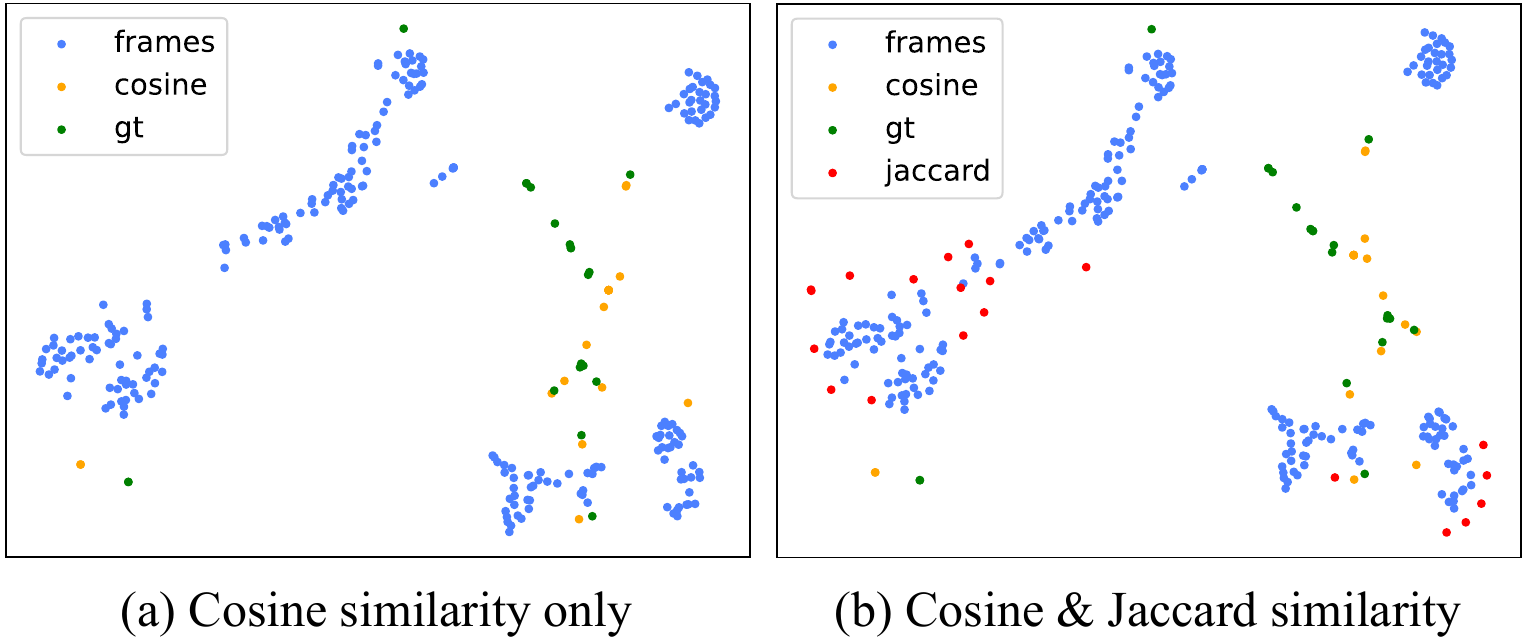}
  \caption{t-SNE visualization of the semantic distribution of video frames, ground truth, and sentences retrieved using cosine similarity and Jaccard similarity.}
  \label{fig: distribution}
\end{figure}

A key challenge in employing the semantic group decoding strategy is constructing a training semantic group whose distribution closely resembles the temporal relationships of video frames. Simply applying random sampling in the sentence bank introduces excessive irrelevant noise, making it difficult for the model to learn the correlation between the training caption and sentences in the semantic group. Moreover, as shown on the left in Figure \ref{fig: distribution} visualized by t-SNE \cite{maaten-jmlr2009-tsne}, the semantic distribution between the semantic group retrieved by cosine similarity and the actual video content is substantial. This indicates that the retrieved sentences lack sufficient diversity to accurately represent the video content. To address this issue, we introduce the key sentences selection module. This module enhances linguistic diversity by selecting sentences based on the Jaccard similarity of extracted nouns and verbs, ensuring a balance between semantic diversity and subject-motion-level correlation with the training caption, as illustrated on the right in Figure \ref{fig: distribution}. By emphasizing key sentence components, KSS introduces a range of subject-action combinations, all while ensuring semantic consistency. Additionally, we develop the probability sampling supervision module, which dynamically selects sentences from the constructed semantic group to serve as ground truth to generate diverse supervision signals. This approach encourages the model to learn the underlying causal relationships of sentences in the semantic group, rather than simply reconstructing the training caption, thereby enhancing its generalization ability.

We conduct experiments on three widely used video captioning datasets, MSVD \cite{chen-acl2011-msvd}, MSR-VTT \cite{xu-cvpr2016-msrvtt}, and VATEX \cite{wang-iccv2019-vatex}. The results demonstrate that SGCap consistently surpasses existing state-of-the-art zero-shot methods across all evaluation metrics. Notably, our method even achieves performance comparable to several fully supervised models. In addition, ablation studies confirm the effectiveness of our semantic group decoding framework and key modules.

In summary, the main contributions of our work are summarized as follows:
\begin{itemize}
    \item We introduce \textbf{SGCap}, a zero-shot video captioning framework leveraging Semantic Group Decoding (\textbf{SGD}) to utilize multi-frame information and explicitly model inter-frame temporal relationships.
    
    \item We design two key components: the \textbf{Key Sentences Selection (KSS)} module and the Probability Sampling Supervision (\textbf{PSS}) module. These modules jointly construct semantically diverse sentence groups that simulate temporal dynamics and guide the model to capture inter-sentence causal relationships, thereby enhancing its generalization ability to video captioning.
    
    \item Experimental results on three benchmark video captioning datasets demonstrate that our method significantly outperforms state-of-the-art methods and achieves performance comparable to some fully supervised models. Ablation studies further validate the effectiveness of each proposed framework and each key component.
\end{itemize}

\section{Related Work}
\subsection{Video Captioning}
Video Captioning aims to generate a single sentence to describe the video content. With the rapid advancement of deep learning, encoder-decoder based methods have been progressively developed. Early works \cite{venugopalan-iccv2015-s2vt} introduced a sequence-to-sequence framework based on Long Short-Term Memory (LSTM), incorporating a video encoder and a caption decoder. The following works, including \cite{yao-iccv2015-dvets,bin-acmmm-2016bltmvc,li-2022neuro-gcnm}, focused on enhancing the visual encoder by designing architectures to better capture video temporal information or different granularity information. 

Following the advancement of Transformer \cite{vaswani-nips2017-transformer} in computer vision and natural language processing tasks, several works have introduced these architectures to enhance the extraction of diverse visual features or improve the fluency of generated sentences. For instance, Ye \textit{et al.} \cite{ye-cvpr2022-hmn} design a hierarchical attention structure to model relationships at the entity, predicate, and sentence levels. Similarly, Lin \textit{et al.} \cite{lin-cvpr2022-swinbert} introduce a transformer-based end-to-end video captioning model exploiting a sparse attention matrix. More recently, emerging works such as Video-LLaMA \cite{zhang-arxiv2023-videollama} and MA-LMM \cite{he-cvpr2024-mallm} leverage large language models as captioners to describe video content, utilizing the strong generalization capabilities of the vision-language models across diverse scenarios.
    
There are also some works focusing on addressing corner cases in video captioning and exploring different task paradigms. For example, Li \textit{et al.} \cite{li-2023ipm-tfrt} propose a time-frequency recurrent transformer to capture frequency information embedded in long videos for dense video captioning, generating multiple detailed captions based on different video segments. Furthermore, \cite{li-2025pr-fewsupervise} introduce a few-supervise video captioning paradigm, which reduces the number of ground truth captions required in training samples. 

However, these methods rely on large-scale video-text pairs for training, which requires labor-intensive and costly manual annotation of video content. Moreover, few studies have explored the zero-shot video captioning task.

\subsection{Zero-shot Image Captioning}
Vision Language Models (VLMs) \cite{radford-icml2021-clip, yang-cvpr2022-unicl} trained with contrastive loss exhibit impressive performance in many downstream tasks. However, due to the absence of a text decoder and the modality gap between the visual and textual domains, these methods cannot be directly applied to generative tasks, such as image captioning and video captioning.

Prior works \cite{tewel-cvpr2022-zerocap, su-arXiv2022-magic, nukrai-arxiv2022-capdec} employ a pre-trained language model as the decoder and leverage CLIP \cite{radford-icml2021-clip} to compute the cosine similarity between the image and the next text token embedding, generating the sentence recursively. However, computing similarity at each time step disrupts the coherence of the generated sentence and fails to achieve sentence-level semantic alignment with the image.

To efficiently decode image embeddings and improve the coherence of generated sentences, several text-only training methods have been proposed. DeCap \cite{li-iclr2023-decap} and Knight \cite{wang-ijcai2023-knight} train a text decoder to map latent representations from the CLIP text embedding space using only textual data. During the inference stage, image features are projected into the text domain through a cosine similarity-based weighted sum, effectively bridging the modality gap and outperforming prior state-of-the-art methods. Alternatively, C3 \cite{zhang-iclr24-c3} addresses the semantic gap by decomposing it into a constant vector and alignment noise, enabling the text decoder to directly decode image embeddings and generate captions. Other methods, such as IFCap \cite{lee-emnlp2024-ifcap,liu-aaai2024-syntic}, focus on extracting fine-grained information (e.g., entities) from the image or incorporate external pre-trained networks to improve the accuracy of words in the generated sentences.

However, these methods are primarily designed for the image captioning task. While these methods can be adapted to the video captioning task by applying average pooling across all frames, this operation weakens its ability to capture temporal dynamics and may introduce noise, thereby degrading the quality of the generated content.

\section{Method}

\begin{figure*}[t]
  \centering
  \includegraphics[width=\linewidth]{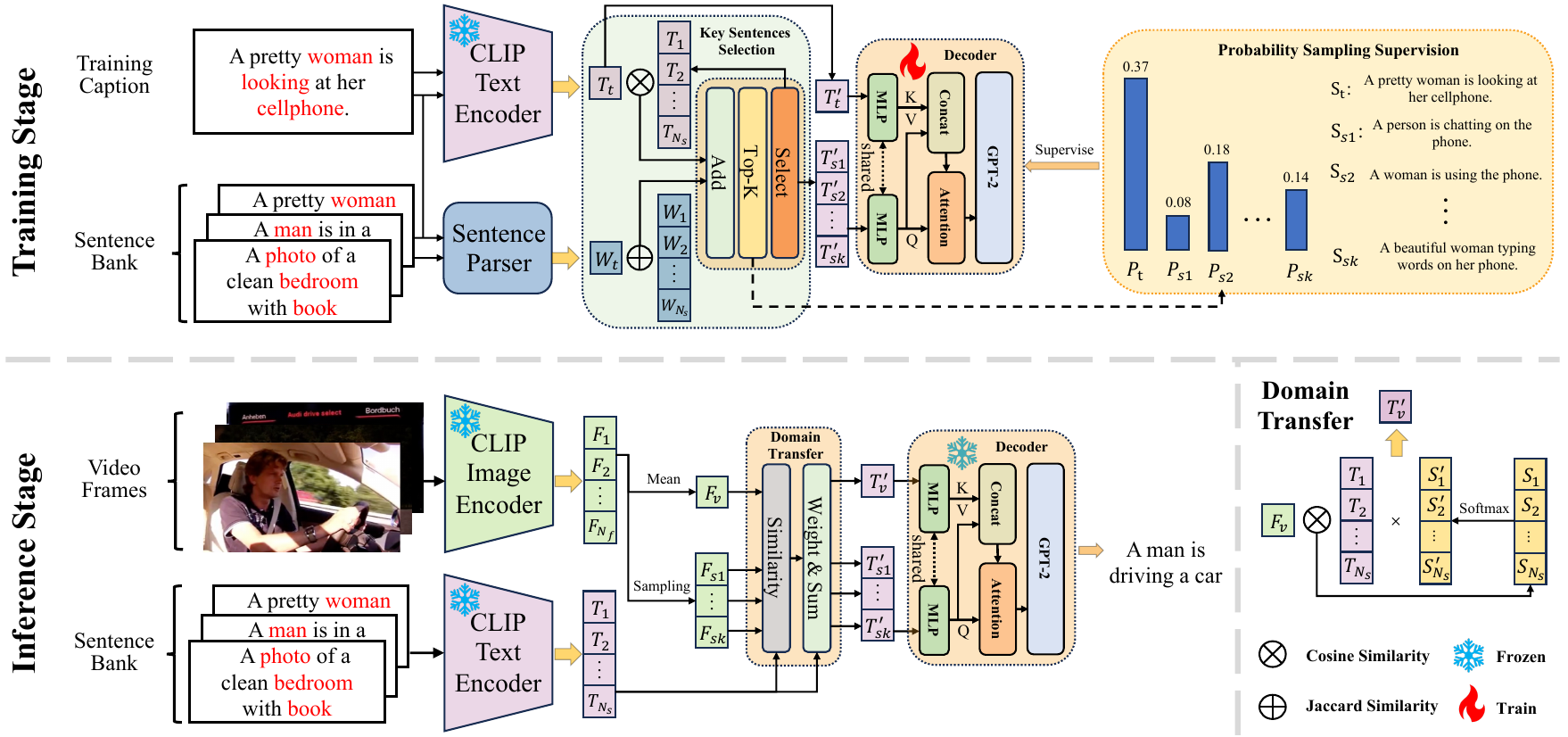}
  \caption{An overview of our proposed framework, SGCap, a zero-shot video captioning framework leveraging \textbf{Semantic Group Decoding (SGD)} strategy to effectively utilize multi-frame information. During the training stage, a semantic group is constructed using the Key Sentences Selection (KSS) module based on cosine similarity and Jaccard similarity scores. The training caption, along with the semantic group, is fed into the fusion module and the language decoder which is supervised by the Probability Sampling Supervision (PSS) module. During the inference stage, frame features are projected into the text domain and decoded by the language decoder to generate captions.}
  \label{fig: method}
\end{figure*}

We propose a novel text-only training model for the zero-shot video captioning task by decoding semantic groups, named SGCap, as illustrated in Figure \ref{fig: method}. 


\subsection{Training Stage}
At this stage, our primary objective is to construct a textual semantic group that reflects the temporal relationships of video frames and train a decoder capable of fusing the training caption with the semantic group. To achieve this goal, we utilize the key sentences selection module and the probability sampling supervision module, ensuring the model learns sufficient linguistic knowledge.

\noindent\textbf{Key Sentences Selection}. This module is designed to retrieve sentences that exhibit high correlation with the training caption while maintaining sufficient semantic diversity. First, the CLIP text encoder is used to encode all the captions into text features, denoted as $T_i, 1 \le i \le N_s$ where $N_s$ represents the number of sentences in the sentence bank. Similarly, sentences are parsed into words using NLTK. Nouns and verbs are extracted to form a set, denoted as $W_i, 1 \le i \le N_s$. Next, cosine similarity and Jaccard similarity are computed between the training caption $T_t$ and all sentences in the sentence bank, as formulated below:

\begin{equation}
    C_{t,i} = \frac{T_t \cdot T_i}{\|T_t\| \, \|T_i\|}, \quad J_{t,i} = \frac{|W_t \cap W_i |}{|W_t \cup W_i |}, 
\end{equation}

\begin{equation}
    P_{t,i} = \sigma \times C_{t,i} + (1-\sigma) \times J_{t,i}, \quad 1 \le i \le N_s 
    \label{equ. prob}
\end{equation}
where $C_{t,i}$, $J_{t,i}$ and $P_{t,i}$ represent the cosine similarity, Jaccard similarity and weighted sum computed between the training caption and the $i$-th sentence in the sentence bank. $\sigma \in (0,1)$ is a hyperparameter controlling the relative importance of the two scores.

After calculating scores for each sentence in the sentence bank, we select the Top-$K$ sentences to construct the semantic group $T_{t,S} = \left\{T_{s_1}, \dots, T_{s_k}\right\}$ which consists of $k$ text vectors. Here, $\left\{s_1, \dots, s_k\right\}$ represents the selected indices in the sentence bank. To enhance the robustness of the retrieval and reconstruction process, we propose a novel element-wise Gaussian noise injection method, which differs from the approach used in Knight \cite{wang-ijcai2023-knight}. For each element in the text vector of the semantic group, the following transformation is applied:

\begin{equation}
T'_{s_i}=T_{s_i}+\epsilon, \quad 1 \le i \le k,
\end{equation}

\begin{equation}
\epsilon\sim\mathcal{N}\left(0,\text{diag}\left(\left\{\frac{1}{N_s}\sum_{i = 1}^{N_s}(T_{ij}-\overline{T_j})^2\right\}_{j = 1}^{d}\right)\right),
\end{equation}
where $\epsilon$ represents the element-wise Gaussian noise. Here, $\overline{T_j}$ is the mean value of the $j$-th feature dimension across all $N_s$ sentences in the sentence bank. The diagonal matrix in the variance of the Gaussian distribution is formed by calculating the variance of each feature dimension $j$ from $1$ to hidden dimension $d$. 

\noindent\textbf{Fusion module}. The fusion module combines the training caption with the semantic group, simulating the fusion process of each video frame. The caption features are first passed through a parameter-shared Feed-Forward Network (FFN) and then fed into a multi-head self-attention layer to obtain the fused feature. After obtaining the fused feature, it is fed into the GPT-2 model to generate captions in an autoregressive manner.

\noindent\textbf{Probability Sampling Supervision}. This module is designed to generate diverse supervision signals for the decoder. By sampling from the training caption and the constructed semantic group, the module encourages the model to learn from the group context rather than directly reconstruct the training caption. The weighted sum calculated in Equation (\ref{equ. prob}) is used for sampling since the probability of a sentence being chosen as the ground truth is correlated with the semantic similarity between the sentence and the training caption. The probability of selecting the training caption is set to $\lambda$, where $\lambda > 0$.

We apply a softmax function to these scores to obtain a normalized probability for each sentence. Then, we sample from these sentences to supervise the generation process using cross-entropy loss, formulated below:

\begin{equation}
    P_t = \left\{\lambda, P_{t,s_1},\dots,P_{t,s_k}\right\},
\end{equation}

\begin{equation}
    \mathcal{L}_{\theta}=-\sum_{i=1}^{k+1}\frac{e^{P_t^i}}{\sum_{i'=1}^{k+1}e^{P_t^{i'}}}\frac{1}{N_w}\sum_{t=1}^{N_w} y_t^i logP\left(\hat{y}_t | \hat{y}_{<t};\theta\right), 
\end{equation}
where $\theta$ denotes the model parameters, and $N_w$ represents the number of words in a sentence. $\hat{y}_t \in \mathbb{R}^v$ and $y_t^i \in \mathbb{R}^v$ refer to the predicted probability distribution of the $t$-th word and the one-hot vector of the $t$-th word in the $i$-th ground truth sentence, respectively. Here, $v$ denotes the size of the vocabulary.

\begin{table*}[!t]
    \begin{center}
    \setlength{\tabcolsep}{4.5pt}
    \caption{The comparison with state-of-the-art methods. All methods are trained on the sentence bank that utilizes the ground truth sentences from each respective dataset. “Pretrained” indicates that the pre-trained GPT-2 checkpoint is utilized, and “*” donotes the method adapted from the zero-shot image captioning task, while “\dag” denotes the method uses video-text pairs for fully supervised training. Highest scores are marked in blodface and second-best scores are underlined.}
    \label{Tab: quantitive_msrvtt_msvd}
    \begin{tabular}{lccccccccccccc}
        \toprule[0.75pt]
         \multirow{2}{*}{\textbf{Method}}&  \multicolumn{6}{c}{MSR-VTT} & &\multicolumn{6}{c}{MSVD} \\
         \cmidrule[0.5pt](l){2-7} \cmidrule[0.5pt](r){9-14}
                                &   B@1$\uparrow$& B@4$\uparrow$& M$\uparrow$& R-L$\uparrow$& C$\uparrow$& S$\uparrow$   &   &  B@1$\uparrow$& B@4$\uparrow$& M$\uparrow$& R-L$\uparrow$& C$\uparrow$& S$\uparrow$\\
         \toprule[0.5pt]
         \multicolumn{14}{c}{fully supervised training} \\
         \toprule[0.5pt]
         SGN$^\dag$(\cite{ryu-aaai2021-sgn}) & - & 40.8 & 28.3 & 60.8 & 49.5 & - & &- & 52.8 & 35.5 & 72.9 & 94.3 & - \\
         HMN$^\dag$(\cite{ye-cvpr2022-hmn}) & - & 43.5 & 29.0 & 62.7 & 51.5 & - & &- & 59.2 & 37.7 & 75.1 & 104.0 & - \\
         CoCap$^\dag$(\cite{shen-iccv2023-cocap}) & - & 44.4 & 30.3 & 63.4 & 57.2 & - && - & 60.1 & 41.4 & 78.2 & 121.5 & - \\
         \toprule[0.5pt]
         \multicolumn{14}{c}{text-only training} \\
         \toprule[0.5pt]
         ZeroCap*(\cite{tewel-cvpr2022-zerocap}) & - & 2.3 & 12.9 & 30.4 & 5.8 & - & &- & 2.9 & 16.3 & 35.4 & 9.6 & - \\
         MAGIC*(\cite{su-arXiv2022-magic}) & 22.3 & 5.5 & 13.3 & \underline{35.4} & 7.4 & 4.2 & & 24.7 & 6.6 & 16.1 & 40.1 & 14.0 & 2.9 \\
         CapDec*(\cite{nukrai-arxiv2022-capdec}) & 30.2 & 8.9 & 23.7 & 17.2 & 11.3 & 5.9 & & 33.1 & 7.9 & 23.3 &25.2 & 34.5 & 3.2 \\
         Knight(\cite{wang-ijcai2023-knight}) & 72.6 & 25.4 & \textbf{28.0} & 50.7 & 31.9 & \textbf{8.5} & & 73.1 & 37.7 & 36.1 & 66.0 & 63.8 & 5.0 \\ 
         DeCap*(\cite{li-iclr2023-decap}) & - & 23.1 & 23.6 & - & 34.8 & - & & - & - & - & - & - & - \\
         IFCap*(\cite{lee-emnlp2024-ifcap}) & - & 27.1 & 25.9 & - & 38.9 & 6.7 & &- & 40.6 & 34.2 & - & 83.9 & 6.3 \\
         \toprule[0.5pt]
         Ours  &  \underline{79.8} & \underline{32.6} & 26.4 & \textbf{55.3} & \underline{42.4} & 7.1 & & \underline{80.8} & \underline{49.3} & \underline{37.8} & \textbf{72.2} & \textbf{93.5} & \underline{8.1} \\
         Ours(Pretrained) & \textbf{80.2} & \textbf{33.3} & \underline{26.6} & \textbf{55.3} & \textbf{43.4} & \underline{7.3} & & \textbf{80.9} & \textbf{50.1} & \textbf{38.0} & \underline{71.9} & \underline{91.5} & \textbf{8.3} \\
         \toprule[0.75pt]
    \end{tabular}
    \end{center}
    \vspace{-5mm}
\end{table*}

\subsection{Inference Stage}
In this stage, the objective is to decode multi-frame information and generate a descriptive sentence. To bridge the gap between the visual domain and the text domain, we employ a domain transfer module to retrieve text features, as done in previous zero-shot image captioning works. The video frames are first passed through a CLIP image encoder to extract the visual features. To maintain the same setup as the training stage, which includes a training caption and a semantic group, we perform average pooling over all frame features to obtain a global video content vector $F_v$. Next, we uniformly sample $k$ frames to form the semantic group $\left\{ F_{s_1}, \dots ,F_{s_k} \right\}$. 

\noindent\textbf{Domain Transfer}. The domain transfer module is employed to map the each frame feature and the global video content vector into the text domain. We use the global video content as an example. First, cosine similarity is calculated to obtain the similarity score with all sentences embeddings in the sentence bank, resulting in $S_v=\left\{S_{v,1}, \dots ,S_{v,N_s}\right\}$, where $N_s$ represents the number of sentences in the sentence bank. A Softmax function is applied to normalize the scores, obtaining a weight for each text feature. Finally, we apply a weighted sum to all text features to acquire the vector that represents the semantics of the frame, which avoids the domain gap by training and inferring in the same CLIP text domain. The calculation process is expressed as follows:

\begin{equation}
    S_{v, i} = \frac{F_v \cdot T_i}{\|F_v\| \, \|T_i\|}, \quad 1 \le i \le N_s,  
\end{equation}

\begin{equation}
    T'_v = \sum_{i=1}^{N_s}\frac{e^{(S_{v,i} / \tau)}}{\sum_{i'=1}^{N_s} e^{(S_{v,i'} / \tau)}}T_i,
\end{equation}
where $T'_v$ is the global video content vector mapped into the text domain and $\tau > 0$ is a temperature hyperparameter that controls the smoothness of the distribution generated by the softmax function.

After obtaining the retrieved text and the semantic group, these features are fed into the fusion module and the language decoder. During the generation process, we use beam search to enhance the coherence of the decoded sentence.

\section{Experiments}
\subsection{Experiment Setup}

\begin{table}[!t]
    \centering
    \setlength{\tabcolsep}{3pt}
    \caption{The quantitative results on VATEX dataset. “\dag” denotes methods trained on video-text pairs, “*” denotes the method adapted from the zero-shot image captioning task.}
    \label{Tab: quantitative_vatex}
    \begin{tabular}{lcccccc}
        \toprule[0.75pt]
         \multirow{2}{*}{Method}&  \multicolumn{6}{c}{VATEX} \\
         \cmidrule[0.5pt]{2-7}
            &   B@1$\uparrow$& B@4$\uparrow$& M$\uparrow$& R-L$\uparrow$& C$\uparrow$& S$\uparrow$ \\
        \toprule[0.5pt]
        \multicolumn{7}{c}{fully supervised training} \\ 
        \toprule[0.5pt]
        VATEX$^\dag$(\cite{wang-iccv2019-vatex}) & - & 28.4 & 21.7 & 47.0 & 45.1 & - \\
        SwinBERT$^\dag$(\cite{lin-cvpr2022-swinbert}) & - & 38.7 & 26.2 & 53.2 & 73.0 & - \\
        CoCap$^\dag$(\cite{shen-iccv2023-cocap}) & - & 31.4 & 23.2 & 49.4 & 52.7 & - \\ 
        \toprule[0.5pt] 
        \multicolumn{7}{c}{text-only training} \\ 
        \toprule[0.5pt]
        Knight(\cite{wang-ijcai2023-knight}) & \underline{63.8} & \underline{19.0} & 20.3 & 37.5 & 27.7 & \underline{9.2}\\ 
        DeCap*(\cite{li-iclr2023-decap}) & 60.3 & 15.6 & \underline{20.6} & 34.1 & 27.0 & 9.1\\
        IFCap*(\cite{lee-emnlp2024-ifcap}) & 57.7 & 17.7 & 18.1 & \underline{40.1} & \textbf{32.5} & 7.6\\
         \toprule[0.5pt]
        Ours & \textbf{69.2} & \textbf{25.8} & \textbf{21.4} & \textbf{44.3} & \underline{30.4} & \textbf{9.5} \\
        \toprule[0.75pt]
    \end{tabular}
\end{table}

\begin{table*}[t]
    \centering
    \caption{Ablation studies of the key components of the SGCap. \checkmark indicates the component is enabled.}  \vspace{-3mm}
    \label{Tab: component}
    \begin{tabular}{cccccccccccccccc}
        \toprule[0.75pt]
         \multirow{2}{*}{SGD} & \multirow{2}{*}{PSS} & \multirow{2}{*}{KSS} & \multicolumn{6}{c}{MSR-VTT} & &\multicolumn{6}{c}{MSVD}\\
                   \cmidrule[0.5pt]{4-9}\cmidrule[0.5pt]{11-16}
          &  &  &   B@1$\uparrow$& B@4$\uparrow$& M$\uparrow$& R-L$\uparrow$& C$\uparrow$& S$\uparrow$ & & B@1$\uparrow$& B@4$\uparrow$& M$\uparrow$& R-L$\uparrow$& C$\uparrow$& S$\uparrow$ \\
          \toprule[0.5pt]
          \checkmark &  &  & 77.4 & 27.8 & 25.3 & 52.5 & 39.7 & 7.0 & & 78.7 & 46.4 & 37.1 & 70.6 & 85.0 & 8.0\\
          \checkmark & \checkmark &  & 78.1 & 31.1 & 25.7 & 54.1 & 40.7 & 7.0 & & 80.1 & 48.1 & 37.5 & 71.5  & 85.5 & 8.1 \\
          \checkmark &  &\checkmark  & 78.0 & 28.2 & 25.3 & 52.7 & 41.0 & 7.1 & & 79.7 & 46.8 & 37.1 & 70.5 & 89.1 & 8.2\\
          \checkmark & \checkmark & \checkmark & \textbf{80.2} & \textbf{33.3} & \textbf{26.6} & \textbf{55.3} & \textbf{43.4} & \textbf{7.3} & & \textbf{80.9} & \textbf{50.1} & \textbf{38.0} & \textbf{71.9} & \textbf{91.5} & \textbf{8.3}\\
         \toprule[0.75pt]
    \end{tabular}
\end{table*}

\begin{table*}[t]
    \centering
    \caption{Ablation studies of the different numbers of sentences in training and frames in inference.} \vspace{-3mm}
    \begin{tabular}{cccccccccccccc}
        \toprule[0.75pt]
         \multirow{2}{*}{$K$}&  \multicolumn{6}{c}{MSR-VTT} & &\multicolumn{6}{c}{MSVD} \\
         \cmidrule[0.5pt]{2-7} \cmidrule[0.5pt]{9-14}
                                &   B@1$\uparrow$& B@4$\uparrow$& M$\uparrow$& R-L$\uparrow$& C$\uparrow$& S$\uparrow$    &  &  B@1$\uparrow$& B@4$\uparrow$& M$\uparrow$& R-L$\uparrow$& C$\uparrow$& S$\uparrow$\\
        \toprule[0.5pt]
        1 & 77.0 & 28.9 & 25.2 & 52.7 & 39.5 & 7.0 & & 78.6 & 47.1 & 37.0 & 70.6 & 88.2 & 7.9\\
        3 & 79.1 & 31.8 & 26.2 & 54.7 & 42.5 & 7.0 & & 79.8 & 48.7 & \textbf{38.1} & 71.4 & 87.9 & 8.2\\
        5 & \textbf{80.2} & \textbf{33.3} & \textbf{26.6} & \textbf{55.3} & \textbf{43.4} & \textbf{7.3} & & \textbf{80.9} & \textbf{50.1} & 38.0 & \textbf{71.9} & \textbf{91.5} & \textbf{8.3} \\ 
        10 & 79.3 & 32.4 & \textbf{26.6} & \textbf{55.3} & 42.9 & 7.2 & &  80.7 & 49.3 & 38.0 & 71.8 & 88.6 & 8.1\\
        20 & 77.7 & 31.2 & 26.1 & 54.5 & 39.9 & 7.0 & & 80.1 & 49.5 & \textbf{38.1} & \textbf{71.9} & 89.4 & 8.0\\
         \toprule[0.75pt]
    \label{Tab: number of frames}
    \end{tabular}
    \vspace{-5mm}
\end{table*}

\noindent\textbf{Datasets}. For the video captioning task, we conduct experiments on three widely used benchmarks: MSVD \cite{chen-acl2011-msvd}, MSR-VTT \cite{xu-cvpr2016-msrvtt}, and VATEX \cite{wang-iccv2019-vatex}. We use the sentences in each dataset as the sentence bank and evaluate all methods on the test split, as done in other video captioning and zero-shot tasks.

\noindent\textbf{Metrics}. We follow standard metrics in image and video captioning tasks, namely BLEU (B)\cite{papineni-acl2002-bleu}, METEOR (M)\cite{denkowski-smt2014-meteor}, ROUGE-L(R-L)\cite{lin-acl2004-rouge}, CIDEr-D (C)\cite{vedantam-cvpr2015-cider}, and SPICE (S)\cite{niu-tip2022-spice}. 

\noindent\textbf{Implementation Details}. We use NLTK toolkit to extract verbs and nouns in sentences. For the CLIP\cite{radford-icml2021-clip} model, we choose the ViT-B/32 architecture as our baseline rather than larger versions to reduce computational cost, which encodes both text and image as a 512-dimensional vector. Following prior works\cite{li-iclr2023-decap, lee-emnlp2024-ifcap}, we choose the standard version of GPT-2\cite{radford-openai2019-gpt2} to decode groups. The intermediate size for the feed-forward network and linear layers is set to 4096. For the best version of our model, we set hyper-parameters $\sigma = 0.5$, $\lambda=1$, $\tau=0.01$. We optimize the model with the AdamW \cite{Loshchilov-iclr2019-adamw} optimizer with a learning rate of 1e-4. The global batch size is set to 256. An early stopping strategy is employed. The training process on the MSR-VTT dataset takes less than half an hour, and the training process on the MSVD dataset takes less than 15 minutes using 1 NVIDIA RTX-4090 GPU. During inference, we use the beam search with beam size 5.

\subsection{Quantitative results}
In this section, we compare the test results of our method with other state-of-the-art zero-shot captioning models. It can be observed from Table \ref{Tab: quantitive_msrvtt_msvd} that our method surpasses other state-of-the-art zero-shot approaches on both the MSVD and MSR-VTT datasets and achieves performance comparable to some fully supervised methods. Our model even outperforms the fully supervised methods SGN \cite{ryu-aaai2021-sgn} and HMN \cite{ye-cvpr2022-hmn} on the MSVD dataset in terms of the METEOR metric, demonstrating the effectiveness of our approach. 

For the VATEX \cite{wang-iccv2019-vatex} dataset, we present the comparison results in Table \ref{Tab: quantitative_vatex}. The performance of our model outperforms other state-of-the-art zero-shot methods, while it lags behind models trained on video-text pairs. This discrepancy is attributed to the higher caption complexity in the VATEX dataset and the reduced accuracy of CLIP when retrieving long sentences.

\subsection{Ablation Study}
\noindent\textbf{Key Components}. 
We evaluate the performance of each component by detaching the modules from the complete model, as illustrated in Table \ref{Tab: component}. It can be observed the PSS module significantly improves the n-gram accuracy in terms of the B@4 metric and provides improvements on other metrics. Similarly, the evaluation scores of all metrics benefit from the KSS module. SGCap achieves the best performance when all key components are enabled.

\noindent\textbf{Number of Frames}. We aim to cover as many scenarios as possible with the sampled frames to represent the global semantics. However, using more frames to generate captions introduces additional retrieval noise when building the semantic group during the training stage. As shown in Table \ref{Tab: number of frames}, the model achieves the best result when $K$ is set to 5. 

\begin{table}[h]
    \centering
    \setlength{\tabcolsep}{4pt}
    \caption{Ablation studies of different noise types. $\sigma_s$ represents the mean value of the standard deviation of the text vectors in the sentence bank.}
    \label{Tab: noise}
    \begin{tabular}{ccccccc}
        \toprule[0.5pt]
         \multirow{2}{*}{Noise}&  \multicolumn{6}{c}{MSR-VTT} \\
         \cmidrule[0.5pt]{2-7}
            &   B@1$\uparrow$& B@4$\uparrow$& M$\uparrow$& R-L$\uparrow$& C$\uparrow$& S$\uparrow$ \\
        \toprule[0.5pt]
        0 & 76.8 & 30.0 & 23.9 & 51.3 & 39.9 & 6.5\\
        $\mathcal{N}\sim(0,1)$ & 69.9 & 22.9 & 21.4 & 46.5 & 25.9 & 5.1\\
        $\mathcal{N}\sim(0,\sigma^2_s)$ & 78.0 & 31.4 & 25.8 & 54.3 & 41.3 & 7.0\\
        Ours & \textbf{80.2} & \textbf{33.3} & \textbf{26.6} & \textbf{55.3} & \textbf{43.4} & \textbf{7.3}\\
         \toprule[0.75pt]
    \end{tabular}
\end{table}

\begin{table*}[t]
    \centering
    \caption{Ablation studies of different $\sigma$.} 
    \label{Tab: sigma}
    \begin{tabular}{cccccccccccccc}
        \toprule[0.75pt]
         \multirow{2}{*}{$\sigma$}&  \multicolumn{6}{c}{MSR-VTT} & &\multicolumn{6}{c}{MSVD} \\
         \cmidrule[0.5pt]{2-7} \cmidrule[0.5pt]{9-14}
                                &   B@1$\uparrow$& B@4$\uparrow$& M$\uparrow$& R-L$\uparrow$& C$\uparrow$& S$\uparrow$    &  &  B@1$\uparrow$& B@4$\uparrow$& M$\uparrow$& R-L$\uparrow$& C$\uparrow$& S$\uparrow$\\
         \toprule[0.5pt]
         0 & 79.7 & 31.5 & 26.5 & 55.0 & 43.0 & 7.2 & & 79.3 & 48.5 & 38.0 & 71.9 & 89.4 & 8.2\\
         0.3 & 79.5 & 32.5 & \textbf{26.6} & 55.1 & \textbf{43.7} & \textbf{7.3} & & 80.3 & 49.6 & \textbf{38.2} & \textbf{72.4} & 89.5 & 8.3\\
         0.5 & \textbf{80.2} & \textbf{33.3} & \textbf{26.6} & \textbf{55.3} & 43.4 & \textbf{7.3} & & \textbf{80.9} & \textbf{50.1} & 38.0 & 71.9 & \textbf{91.5} & \textbf{8.3}\\
         0.7 & 79.3 & 32.6 & \textbf{26.6} & \textbf{55.3} & 42.7 & 7.2 & & 80.4 & 50.0 & 38.0 & 72.1 & 90.8 & 8.2\\
         1.0 & 78.1 & 31.1 & 25.7 & 54.1 & 40.7 & 7.0 & & 79.9 & 48.7 & 37.7 & 71.3 & 88.9 & 8.0\\
    \toprule[0.75pt]
    \end{tabular}
\end{table*}

\begin{table*}[!t]
    \centering
    \caption{Ablation studies of different $\lambda$.}
    \label{Tab: lambda}
    \begin{tabular}{cccccccccccccc}
        \toprule[0.75pt]
         \multirow{2}{*}{$\lambda$}&  \multicolumn{6}{c}{MSR-VTT} & & \multicolumn{6}{c}{MSVD} \\
         \cmidrule[0.5pt]{2-7} \cmidrule[0.5pt]{9-14}
                                &   B@1$\uparrow$& B@4$\uparrow$& M$\uparrow$& R-L$\uparrow$& C$\uparrow$& S$\uparrow$    &  &  B@1$\uparrow$& B@4$\uparrow$& M$\uparrow$& R-L$\uparrow$& C$\uparrow$& S$\uparrow$\\
         \toprule[0.5pt]
         0.5 & 78.9 & 32.3 & 26.3 & 55.0 & 42.0 & 7.0 & & 80.5 & 50.5 & 37.7 & 71.8 & 88.9 & 8.1\\
         1.0 & \textbf{80.2} & \textbf{33.3} & \textbf{26.6} & \textbf{55.3} & 43.4 & \textbf{7.3} & & \textbf{80.9} & 50.1 & 38.0 & 71.9 & 91.5 & \textbf{8.3}\\
         1.5 & 79.4 & 32,2 & 26.2 & 55.0 & 43.3 & 7.2 & & 80.7 & \textbf{50.7} & \textbf{38.6} & \textbf{72.5} & \textbf{92.5} & 8.1\\ 
         2.0 & 79.5 & 32.7 & 26.5 & 55.2 & \textbf{43.6} & 7.1 & & 80.7 & 49.7 & 38.0 & 72.0 & 90.2 & 8.2 \\
    \toprule[0.75pt]
    \end{tabular}
\end{table*}

\begin{figure}[t]
    \centering
    \includegraphics[width=\linewidth]{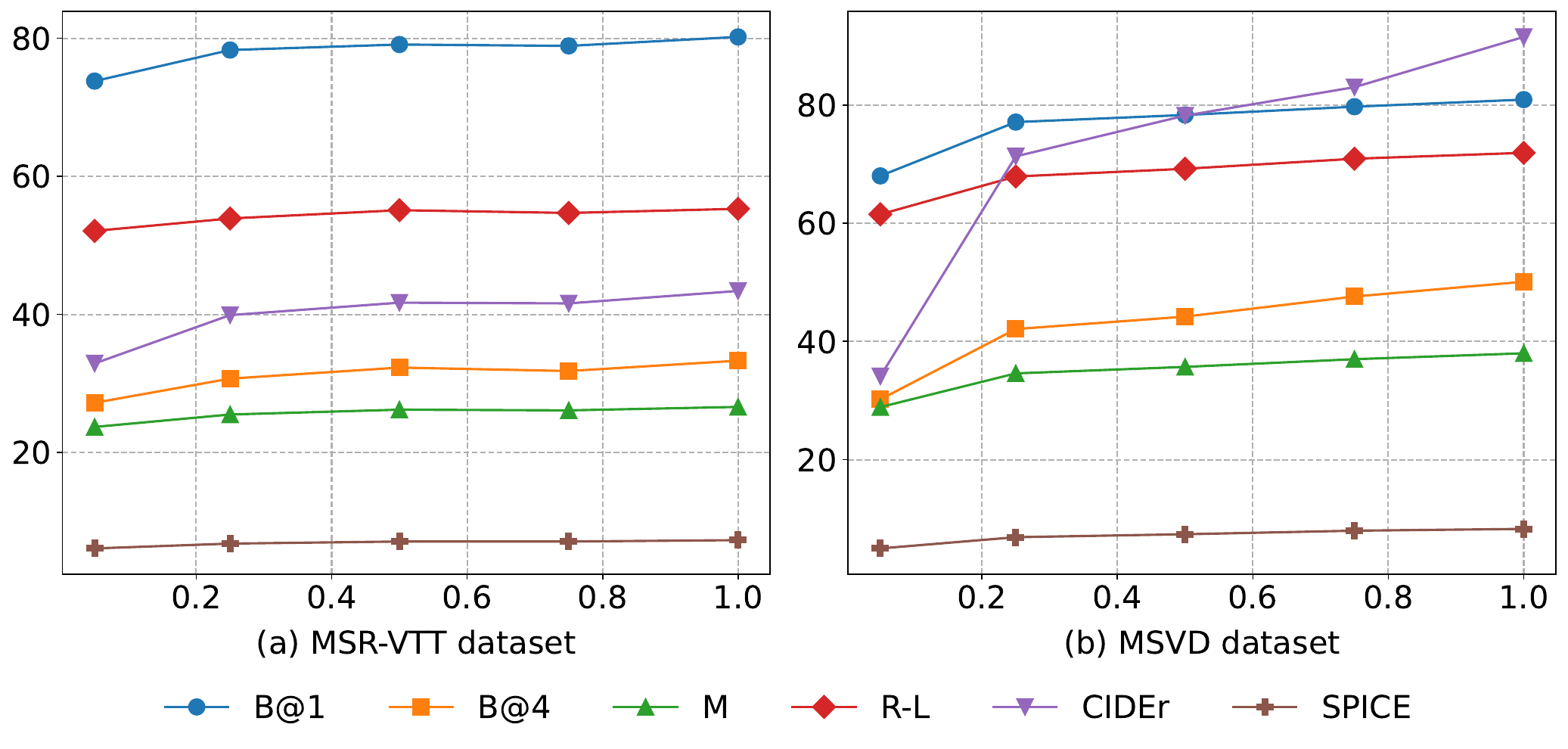}
    \caption{The ablation results of our method under different sentence bank size on MSR-VTT dataset (left) and MSVD dataset (right)}
    \label{fig: ablation_size}
\end{figure}

\begin{figure}[!t]
    \includegraphics[width=\linewidth]{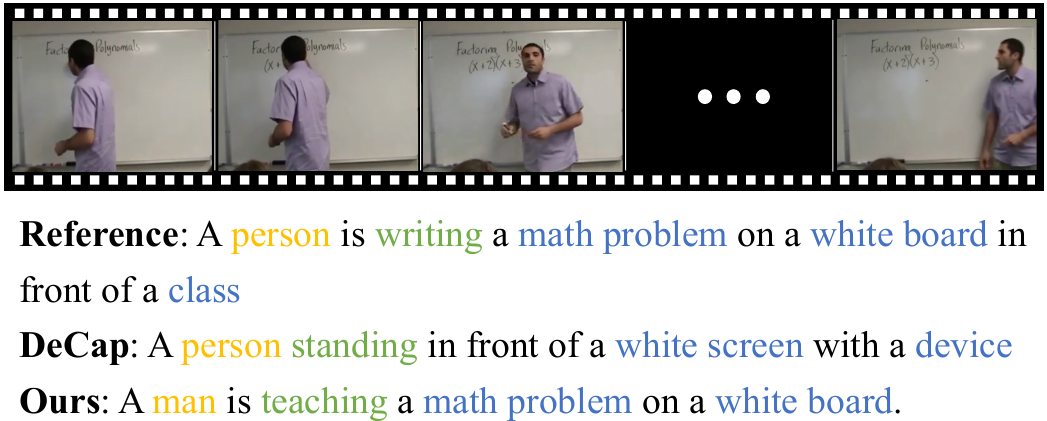}
    \includegraphics[width=\linewidth]{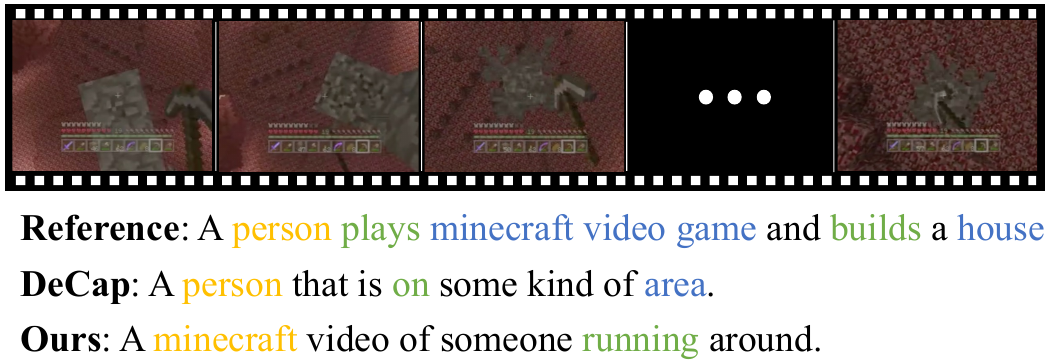}
    \includegraphics[width=\linewidth]{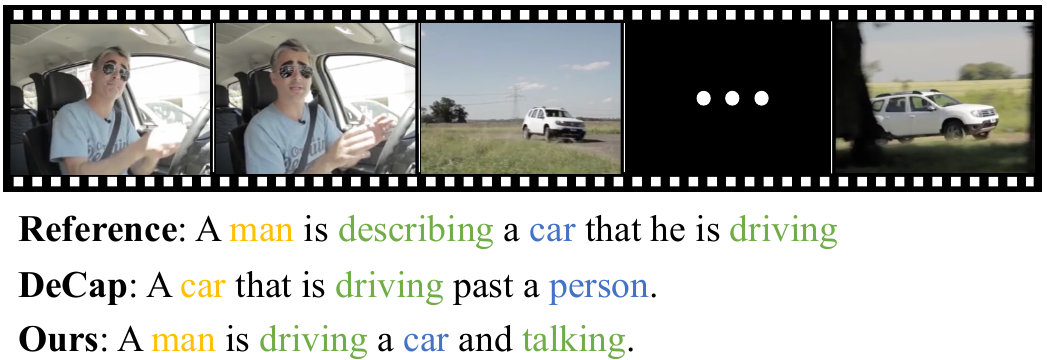}
    \caption{The comparison of qualitative results on the MSR-VTT dataset between our method and DeCap.}
    \label{fig: qualitative}
\end{figure}

\noindent\textbf{Sentence Bank Size}. We experiment with different sizes of the sentence bank to explore its influence on performance. We randomly select parts of ground truth sentences for training. The results are shown in Figure \ref{fig: ablation_size}. It can be observed that the performance of the model improves as the size of the sentence bank increases from 5\% to 100\%, achieving the best result when all captions are utilized for training. However, it is worth noting that SGCap does not suffer from a catastrophic performance drop, even with just 5\% of the training sentences. Compared to other image-to-video adaption models, SGCap still has a significant improvement.

\noindent\textbf{Types of Noise}. We conduct experiments with different noise types in Table \ref{Tab: noise} to investigate the distribution characteristics of elements in the content vector. The second row shows the standard Gaussian noise. In the third row, we calculate the standard deviation of all sentences, resulting in a scalar as standard deviation to generate the Gaussian noise. In the last row, we employ our element-wise Gaussian noise. The results demonstrate the effectiveness of our element-wise noise.


\noindent\textbf{Hyper-parameter $\sigma$ and $\lambda$}. The former $\sigma$ is used to balance the importance of the cosine similarity and the Jaccard similarity scores. As shown in Table \ref{Tab: sigma}, the model achieves the best performance when $\sigma$ is set to 0.5, which demonstrates the effectiveness of both cosine similarity and Jaccard similarity. $\lambda$ is a hyperparameter that determines the probability of the training captions sampled by PSS module. As shown in Table \ref{Tab: lambda}, the model achieves the best performance on the MSR-VTT dataset when $\lambda=1$, and on the MSVD dataset when $\lambda=1.5$ or $\lambda=1.0$.

\subsection{Qualitative Results}
We compare the performance of DeCap and our proposed SGCap on the MSR-VTT dataset. As demonstrated in Figure \ref{fig: qualitative}, in the first video clip, DeCap only identifies the motion “standing”, which fails to capture the actual action. In contrast, SGCap interprets the video context more accurately, generating the correct motion “teaching”. Similarly, in the second scenario, DeCap detects the subjects “car” and “man” but struggles to produce coherent captions due to the application of average pooling across all frames. This confuses the decoder, leading to causally incorrect sentences. Our method, however, correctly outputs the action “a man is driving a car and talking.”

However, when handling uncommon scenarios, both DeCap and SGCap fail to accurately capture subjects and motions, and generate the corresponding nouns and verbs, as shown in the third video.

\section{Conclusion}
This work presents a zero-shot video captioning method called SGCap based on the semantic group decoding strategy to address the problem of training the model without video-text pairs. The key sentence selection and probability sampling supervision modules are employed to ensure the model learns accurate and diverse linguistic knowledge, which enhances the quality of generated captions. Extensive experiments and ablation studies on several benchmarks validate the effectiveness of our approach. In future, we may explore other semantic group construction strategies, while how to adapt other image-based models for video understanding could be further investigated.

\bibliographystyle{plain}
\bibliography{SGCap-draft-arxiv-zeyu}

\cleardoublepage

\section*{\textbf{\Large Appendix}}
\subsection{Alogorithm}
In this section, we summarize the algorithm of our SGCap during the training and inference stage.
\begin{algorithm}
	\caption{SGCap Training Algorithm}
	\label{alg:training}
	\small
	\begin{algorithmic}[1]
		\REQUIRE Training captions $\mathcal{Y} = \{ y_t | 1 \le t \le N_w, y_t \in \mathbb{R}^{N_v} \}$. Fixed CLIP text encoder $f_\mathcal{Y} : \mathcal{Y} \mapsto  \mathbb{R}^d$
		\WHILE{ id $\le$ Epoch }
		\FOR {each batch}
		\STATE Use cosine and Jaccard similarity to retrieve sentences with high correlation for the $t$-th training caption $T_t$, obtaining a semantic group $T_{t,S} = \left[T_{s1}, \dots ,T_{s2}\right]$;
		\STATE Sample from the training caption and the semantic group based on similarity scores to obtain the target ground truth $Y_t$;
		\STATE Feed the training caption and the semantic group into the fusion module and the decoder;
		\STATE Compute loss and perform backpropagation.
		\ENDFOR
		\ENDWHILE
	\end{algorithmic}
\end{algorithm}

\begin{algorithm}
	\caption{SGCap Inference Algorithm}
	\label{alg:inference}
	\small
	\begin{algorithmic}[1]
		\REQUIRE Test video $\mathcal{X} = \{ \mathbf{X}_i \ | \ 1 \le i \le N_f, \ \mathbf{X}_i \in \mathbb{R}^{c\times w \times h} \}$; Fixed CLIP image encoder $f_\mathcal{X} : \mathcal{X} \mapsto  \mathbb{R}^d$
		\STATE Sample video frames and encode them using CLIP image encoder to obtain the semantic group;
		\STATE Average the frame features to obtain the global video content vector;
		\STATE Project the content vector and the semantic group into the textual domain;
		\STATE Fuse the content vector and the semantic group;
		\STATE Decode the fused feature and generate captions;
	\end{algorithmic}
\end{algorithm}

\subsection{Experiments}

\subsubsection{Datasets}
MSVD\footnote{\url{https://www.cs.utexas.edu/users/ml/clamp/videoDescription/}}\cite{chen-acl2011-msvd} is a classical video captioning dataset that contains 1,970 short video clips collected from the YouTube website. Each short video clip is associated with 40 corresponding human-annotated sentences. Additionally, the dataset is split into three subsets, training, evaluation, and test set. The training subset consists of 1,200 clips, while the evaluation set has 100 videos. The last 670 videos are distributed to the test subset. 

MSR-VTT\footnote{\url{https://www.kaggle.com/datasets/vishnutheepb/msrvtt?resource=download}}\cite{xu-cvpr2016-msrvtt} is a large-scale video captioning dataset. The dataset contains 10,000 video clips, amounting to 41.2 hours and 200K video-sentence pairs. Each video clip is associated with 20 manually annotated descriptive sentences. Additionally, MSR-VTT provides category information for the videos, with a total of 20 categories, and includes audio information. The dataset is divided into 6,513 video clips for training, 497 for validation, and 2,990 for testing.

VATEX\footnote{\url{https://eric-xw.github.io/vatex-website/about.html}}\cite{wang-iccv2019-vatex} is a subset of the Kinetics-600\cite{kay-arxiv2017-kinetics} dataset. It consists of 25,991 training videos, 3,000 validation videos, 6,000 publicly available test videos, and 6,278 non-public test videos. This paper uses only the publicly available test videos. Additionally, due to the unavailability of some videos, our experiment utilizes 22,765 training videos, 2,644 validation videos, and 5,244 test videos.

The sentence bank is constructed by utilizing the ground truth sentences from each respective dataset. We evaluate the model performance on the test set of the three datasets following other video captioning methods.

\subsubsection{Metrics}
We follow standard metrics in the image captioning and the video captioning task, namely BLEU (B)\cite{papineni-acl2002-bleu}, METEOR (M)\cite{denkowski-smt2014-meteor}, ROUGE-L(R-L)\cite{lin-acl2004-rouge}, CIDEr-D (C)\cite{vedantam-cvpr2015-cider}, and SPICE (S)\cite{niu-tip2022-spice}. 
Among them, BLEU records the co-occurrences of n-grams between candidate and reference sentences; METEOR aligns words in candidate and reference sentences and minimizes the number of chunks containing contiguous and identically ordered tokens; ROUGE-L (R-L) measures the longest common subsequence (LCS) between candidate and reference sentences, evaluating the fluency and coherence of the generated text by considering the longest sequence of matching words in order; CIDEr-D computes the cosine similarity between candidate and reference sentences by applying Term Frequency-Inverse Document Frequency (TF-IDF) weighting to each n-gram; SPICE evaluates the meaning and relationships conveyed in candidate and reference sentences by parsing them into scene graphs and computing an F-score based on matching semantic propositions, making it more semantic-aware than other metrics.

\subsubsection{Experiment setup}
All experiments were conducted on a machine running Ubuntu 22.04.4 LTS (kernel version 6.5.0-28-generic) with an Intel Core i9-13900 (13th Gen, 32 threads) processor and 64GB of system memory. The system was equipped with an NVIDIA GeForce RTX 4090 GPU with 24GB of VRAM. The software environment included Python 3.13.2, PyTorch 2.6.0, Torchvision 0.21.0, and Transformers 4.48.2, with CUDA support provided by CUDA 12.8 (V12.8.61).

\subsection{Out-of-Domain Learning}
We evaluate the out-of-domain performance of our method by training the fusion module and the decoder on other textual data. Both the training captions and the sentence bank are constructed using the ground truth captions from the specific dataset. For better comparison, we supplement text data from the widely used image captioning dataset COCO \cite{lin-eccv2014-mscoco}. We conduct the evaluation on three benchmarks: MSR-VTT \cite{xu-cvpr2016-msrvtt}, MSVD \cite{chen-acl2011-msvd}, and VATEX \cite{wang-iccv2019-vatex}, as illustrated in Tables \ref{Tab: msrvtt_transfer}, \ref{Tab: msvd_transfer}, and \ref{Tab: vatex_transfer}, respectively. The results show that the model achieves the highest metric scores when trained on paired textual data. The main reason for this is that textual data exhibit biases across different datasets, which leads to lower scores when evaluated using reference-based metrics. Furthermore, during the inference stage, the retrieval module struggles to capture the sentence that precisely represents the semantics of the video frames.

Although the evaluation scores for transferring the model to other target datasets are not ideal, we can observe from the qualitative results, shown in Figure \ref{fig: qualitative}, that it indicates that the model still generates convincing results without excessive hallucination. For example, in the MSVD dataset, only VATEX outputs the incorrect action "fall off". However, when dealing with more complex scenes or sentences in the VATEX dataset, our method, trained on other text datasets, fails to accurately capture the visual semantics or retrieve the corresponding long sentences to describe the given video, leading to errors and hallucinations.

\begin{table}[H]
	\centering
	\setlength{\tabcolsep}{3pt}
	\caption{Experimental results on MSR-VTT by training the text decoder on other datasets.}
	\label{Tab: msrvtt_transfer}
	\begin{tabular}{ccccccc}
		\hline
		\multirow{2}{*}{Sentence Bank}&  \multicolumn{6}{c}{MSR-VTT} \\
		\cmidrule[0.5pt]{2-7}
		&   B@1& B@4& M& R-L& CIDEr& SPICE \\
		\hline
		COCO & 53.8 & 9.9 & 20.2 & 36.7 & 8.1 & 4.9 \\
		VATEX & 53.2 & 12.1 & 22.5 & 36.9 & 9.1 & 6.4 \\
		MSVD & 67.9 & 20.3 & 21.5 & 49.0 & 21.8 & 5.4 \\
		MSR-VTT & \textbf{80.2} & \textbf{33.3} & \textbf{26.6} & \textbf{55.5} & \textbf{43.3} & \textbf{7.2} \\
		\hline
	\end{tabular}
\end{table}

\begin{table}[H]
	\centering
	\setlength{\tabcolsep}{3pt}
	\caption{Experimental results on MSVD by training the text decoder on other datasets.}
	\label{Tab: msvd_transfer}
	\begin{tabular}{ccccccc}
		\hline
		\multirow{2}{*}{Sentence Bank}&  \multicolumn{6}{c}{MSVD} \\
		\cmidrule[0.5pt]{2-7}
		&   B@1& B@4& M& R-L& CIDEr& SPICE \\
		\hline
		COCO & 56.5 & 16.3 & 28.2 & 45.5 & 20.4 & 6.1\\
		VATEX & 52.6 & 16.5 & 28.3 & 43.7 & 19.2 & 6.9 \\
		MSR-VTT & 71.1 & 32.0 & 29.9 & 60.9 & 52.3 & 5.9\\
		MSVD & \textbf{81.0} & \textbf{50.1} & \textbf{38.0} & \textbf{71.9} & \textbf{91.5} & \textbf{8.3} \\
		\hline
	\end{tabular}
\end{table}

\begin{table}[H]
	\centering
	\setlength{\tabcolsep}{3pt}
	\caption{Experimental results on VATEX by training the text decoder on other datasets.}
	\label{Tab: vatex_transfer}
	\begin{tabular}{ccccccc}
		\hline
		\multirow{2}{*}{Sentence Bank}&  \multicolumn{6}{c}{VATEX} \\
		\cmidrule[0.5pt]{2-7}
		&   B@1& B@4& M& R-L& CIDEr& SPICE \\
		\hline
		COCO & 54.5 &12.2 & 15.6 & 36.6 & 10.5 & 5.6 \\
		MSVD & 50.5 & 10.8 & 14.8 & 32.1 & 12.2 & 5.6 \\
		MSR-VTT & 53.1 & 12.2 & 15.0 & 32.3 & 13.1 & 5.6 \\
		VATEX & \textbf{69.2} & \textbf{25.8} & \textbf{21.4} & \textbf{44.3} & \textbf{30.4} & \textbf{9.5} \\
		\hline
	\end{tabular}
\end{table}

\begin{figure}
	\includegraphics[width=\linewidth]{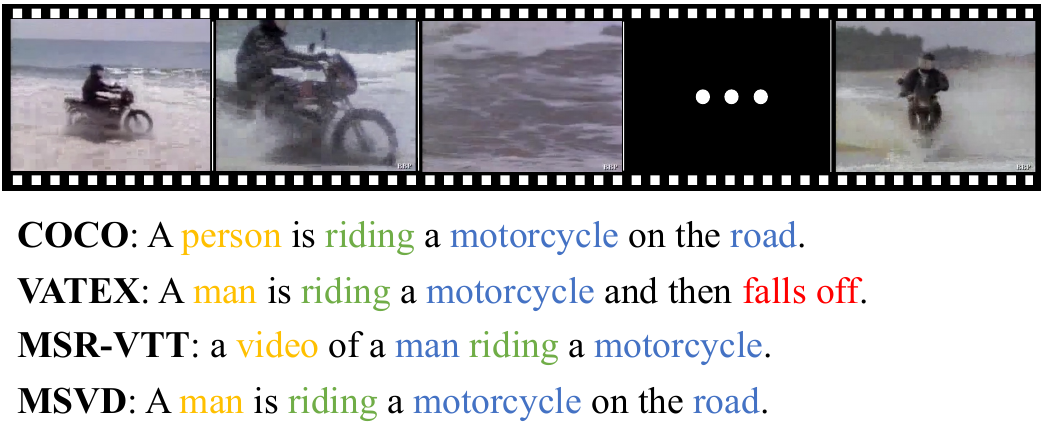}
	\includegraphics[width=\linewidth]{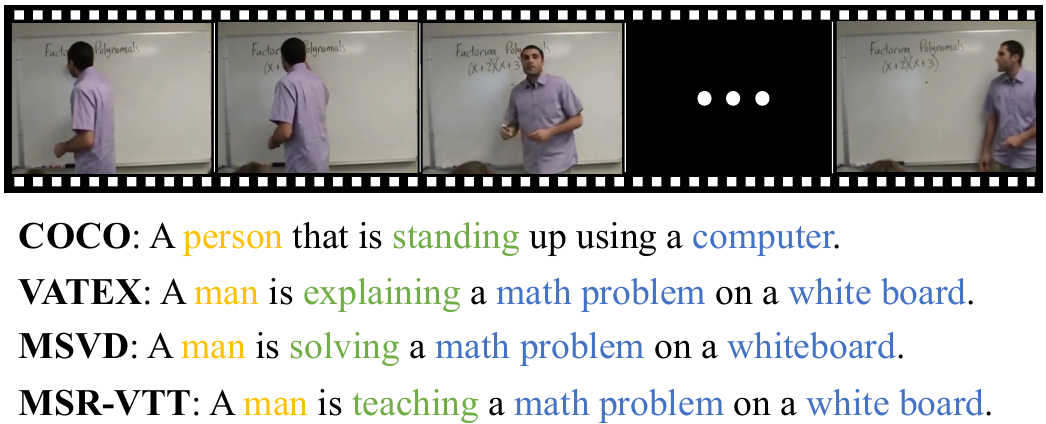}
	\includegraphics[width=\linewidth]{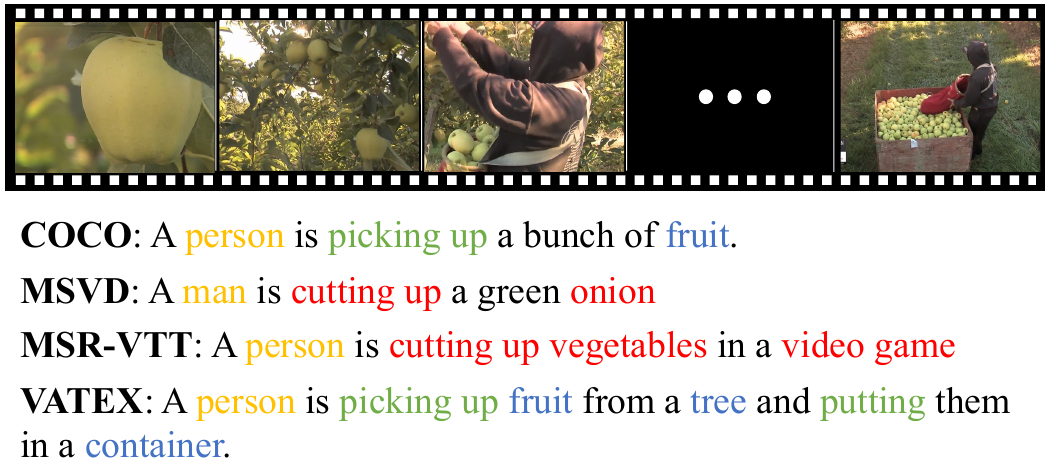}
	\caption{Comparison of qualitative results by transferring the model to the target dataset. The videos are from the test split of the MSVD, MSR-VTT, and VATEX, respectively.}
	\label{fig: qualitative}
\end{figure}

\subsubsection{Temperature $\tau$}
We evaluate the performance under different temperature $\tau$ settings. This hyperparameter balances the sharpness of the normalized distribution, which is crucial when retrieving and building semantic groups. Increasing the temperature smooths the probability distribution, allowing more diverse but possibly irrelevant sentences to be sampled. As shown in Table \ref{Tab: temperature}, it can be observed that the model achieves the best result when the temperature is set to 0.01, striking a balance between sentence diversity and correlation.

\begin{table}[H]
	\centering
	\caption{Experimental results on MSR-VTT with different temperature $\tau$ settings.}
	\label{Tab: temperature}
	\begin{tabular}{ccccccc}
		\hline
		\multirow{2}{*}{$\tau$}&  \multicolumn{6}{c}{MSR-VTT} \\
		\cmidrule[0.5pt]{2-7}
		&   B@1& B@4& M& R-L& CIDEr& SPICE \\
		\hline
		0.001 & 74.0 & 27.6 & 24.3 & 51.0 & 39.2 & 7.0 \\
		0.01 & \textbf{80.2} & \textbf{33.3} & \textbf{26.6} & \textbf{55.5} & \textbf{43.3} & \textbf{7.2} \\
		0.1 & 71.8 & 20.4 & 20.9 & 49.1 & 7.9 & 4.9 \\
		1 & 70.2 & 18.9 & 20.5 & 48.8 & 6.4 & 4.5 \\
		\hline
	\end{tabular}
\end{table}

\subsection{Visualization}
\subsubsection{Key Sentences Selection}
\begin{figure}
	\includegraphics[width=\linewidth]{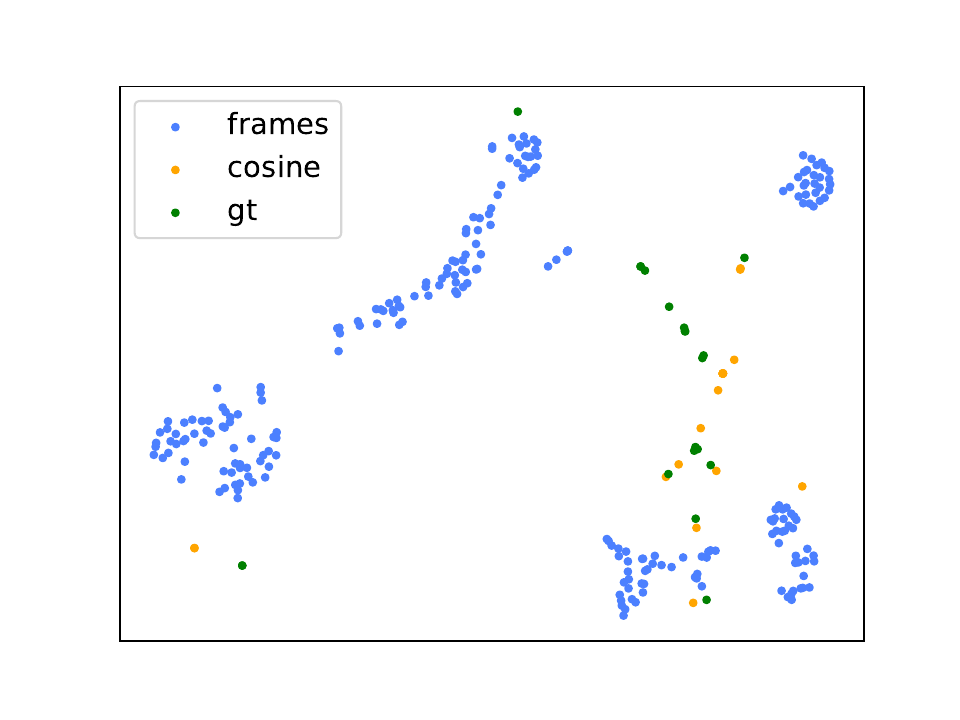}
	\caption{The visualization results of the sentence embeddings. The frame embeddings are in blue, the ground truth sentences are in green, and the sentences retrieved by cosine similarity are in yellow.}
	\label{fig: kss_cosine}
\end{figure}

\begin{figure}
	\includegraphics[width=\linewidth]{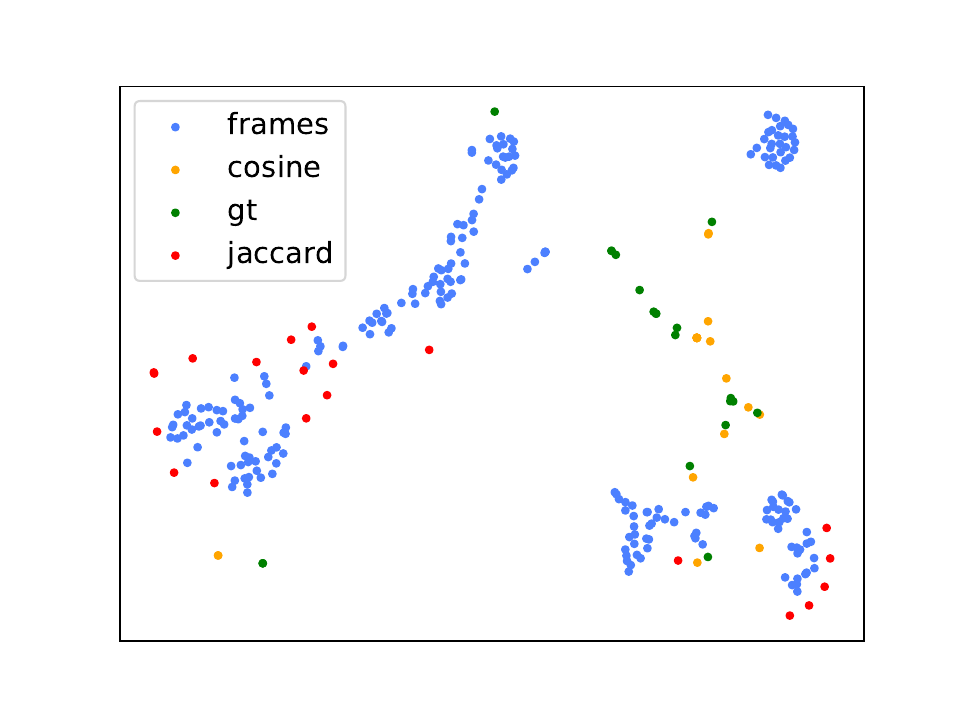}
	\caption{The visualization results of the sentence embeddings. The frame embeddings are in blue, the ground truth sentences are in green, and the sentences retrieved by Jaccard similarity are in red.}
	\label{fig: kss_jaccard}
\end{figure}

In the Key Sentences Selection (KSS) module, we employ the Jaccard similarity to increase the diversity of the semantic group. We visualize the results of the embeddings using only cosine similarity and those combined with Jaccard similarity using the t-SNE \cite{maaten-jmlr2009-tsne} algorithm. As shown in Figure \ref{fig: kss_cosine}, blue, green, and yellow represent the frames, ground truth sentences, and sentences retrieved by cosine similarity. In Figure \ref{fig: kss_jaccard}, red indicates the sentences retrieved using Jaccard similarity. From the figures, we observe that the semantic diversity of sentences retrieved using only cosine similarity is limited, whereas incorporating Jaccard similarity increases diversity while preserving word-level correlation with the ground truth sentences.

\subsubsection{Embedding Variance}
We visualize the variance of each dimension for the text embedding to highlight the importance of element-wise noise in improving the robustness of the retrieval operation. Previous studies like Domino \cite{eyuboglu-iclr2022-domino} and C3 \cite{zhang-iclr24-c3} have summarized the domain gap as the dimensional collapse of the representation space, defined as follows.

\textit{Given a $d$-dimensional representation $\mathbb{R}^d$, define the effective dimension $d_e$ as: } 
\begin{equation}
	d_e = \arg \min_{d^{'}}\frac{\sum_{i=1}^{d^{'}}\sigma_i}{\sum_{i=1}^{d}\sigma_i} \geq \gamma 
\end{equation}
where the $\sigma_i$ are singular values of the representation covariance matrix in decreasing order, and $\gamma$ threshold of the minimum variance explained by the $d_e$ dimension. In other words, the domain gap is primarily represented by a few effective dimensions. 

\begin{figure}
	\centering
	\includegraphics[width=\linewidth]{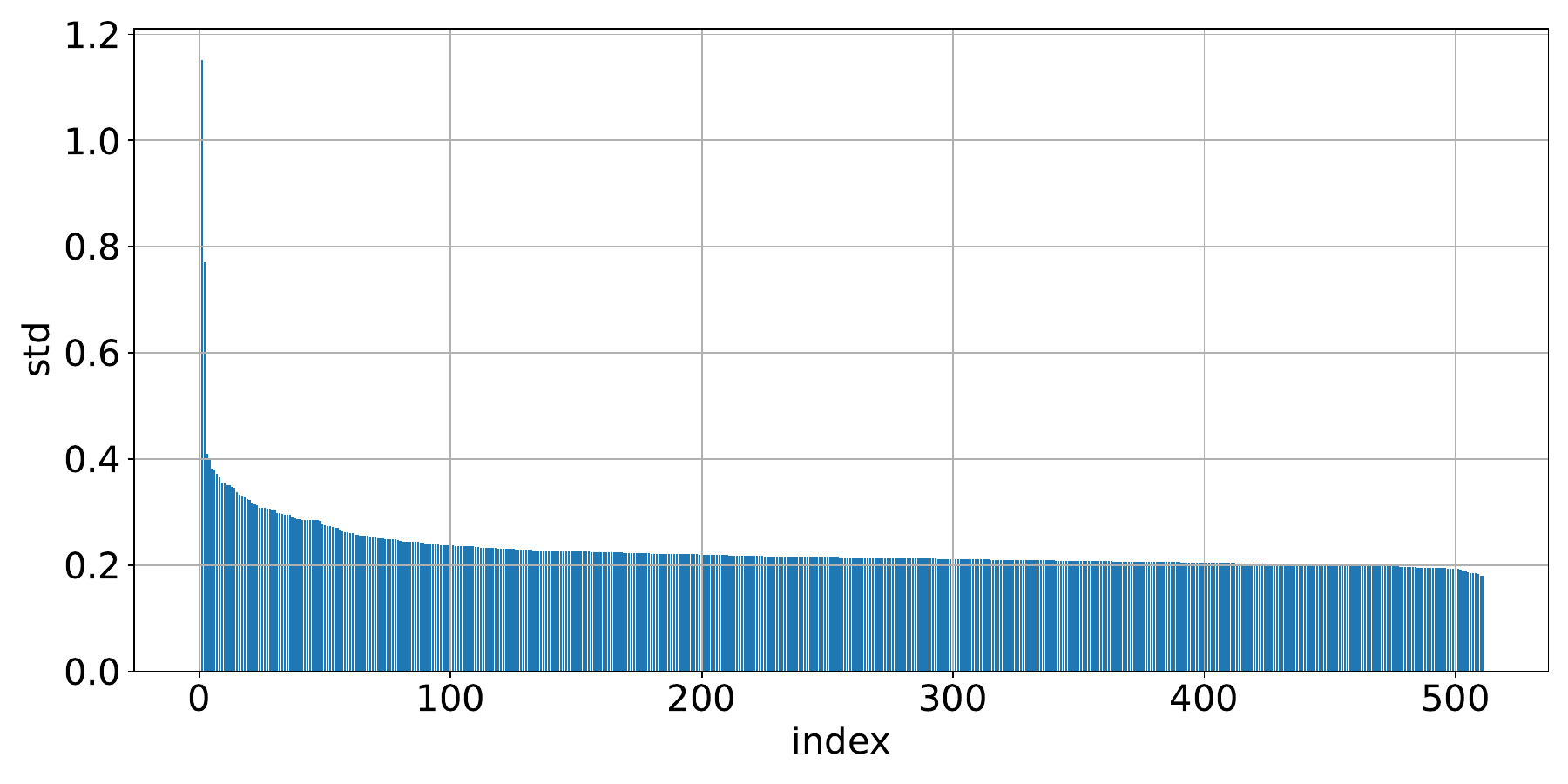}
	\caption{Variance of each dimension for the text embedding encoded by CLIP text encoder.}
	\label{fig: varience}
\end{figure}

As shown in Figure \ref{fig: varience}, we observe that the standard deviation of some dimensions drastically exceeds the average standard deviation, which may indicate the key dimensions. To address this, we adopt element-wise Gaussian noise to maintain the embedding space structure and improve the robustness of the retrieval process.

\subsection{Discussion}
\subsubsection{Probability Sampling Supervision}
The conventional video captioning task trains the model on video-text pairs. Typically, more than 10 sentences are used to describe the video content, ensuring that the model learns from diverse text. The loss function for each video can be formulated as follows:
\begin{equation}
	\mathcal{L}_{\theta} = \frac{1}{N}\sum_{i=1}^N\left[-\sum_{t=1}^{N_w}logP\left(\hat{y}_t | \hat{y}_{<t};\theta\right)\right]
\end{equation}
where $V$ represents the video, and $\theta$ represents the model parameters. $N$ indicates the number of description sentences for each video, and $k$ indicates the $k$-th word in the sentences.

However, for the text-only training scenario, only the specific training caption is utilized to supervise the reconstruction process of the semantic group, causing insufficient learning of diverse linguistic knowledge. This can be formulated as follows:
\begin{equation}
	\mathcal{L}_{\theta} = -\sum_{t=1}^{N_w}logP\left(\hat{y}_t | \hat{y}_{<t};\theta\right)
\end{equation}
where $S$ represents a single sentence embedding.

The Probability Sampling Supervision (PSS) module designs a mechanism to fully exploit the similarity scores and create diverse supervision signals, ensuring the model can also be supervised by other sentences during the backpropagation process. The overall loss function is formulated as follows:
\begin{equation}
	\mathcal{L}_{\theta}=-\sum_{i=1}^{k+1}\frac{e^{P_t^i}}{\sum_{i'=1}^{k+1}e^{P_t^{i'}}}\frac{1}{N_w}\sum_{t=1}^{N_w} y_t^i logP\left(\hat{y}_t | \hat{y}_{<t};\theta\right)
\end{equation}
where $Y_i$ represents the $i$-th sentence, and $k$ indicates the $k$-th step during the autoregressive generation process.

\end{document}